\documentclass[a4paper, 10pt, conference]{ieeeconf}
\IEEEoverridecommandlockouts
\usepackage{epstopdf}
\usepackage{graphics} 
\usepackage{epsfig}
\usepackage{amssymb}  
\usepackage{amsfonts}
\usepackage{amssymb,amsmath}
\usepackage{color}
\usepackage{psfrag}

\renewcommand{\cdot}{}  
\graphicspath{{img/}} 

\newcommand{\executeiffilenewer}[3]{%
  \ifnum\pdfstrcmp{\pdffilemoddate{#1}}%
  {\pdffilemoddate{#2}}>0%
  {\immediate\write18{#3}}\fi%
}

\title{\LARGE \bf Aerial Grasping: Modeling and Control of a Flying Hand%
  \thanks{This work has been funded by the European Commission's
    Seventh Framework Programme as part of the project AIRobots under grant no. 248669.
  }
}

\author{R.T.L.M. Tummers, M. Fumagalli and R. Carloni%
  \thanks{The authors are within the CTIT Institute, Faculty of Electrical Engineering, Mathematics, and Computer Science, University of Twente, The Netherlands. Emails:  r.t.l.m.tummers@alumnus.utwente.nl,
    \{m.fumagalli, r.carloni\}@utwente.nl}
}

\begin{document}

\maketitle

\begin{abstract}
In this paper, we present the design, simulation and experimental validation of a control architecture for a flying hand, i.e., a system made of an unmanned aerial vehicle, a robotic manipulator and a gripper, which is grasping an object fixed on a vertical wall. The goal of this work is to show that the overall control allows the flying hand to approach the wall, to dock on the object by means of the gripper, take the object and fly away. The control strategy has been implemented and validated in the simulated model and in experiments on the complete flying hand system.
\end{abstract}

\section{INTRODUCTION}
Recently the research interest in aerial service robots is
increasing. One of the main goals is to use unmanned aerial vehicles
(UAVs) in real application scenarios to support human beings in all
those activities that require the ability to interact actively and safely with environments not constrained on the ground but airborne \cite{airobots,airobots_video}.

Several works attest the interest in such challenging control
scenarios. For instance, grasping and transportation using a fleet of
quadrotors is considered in \cite{penn}, and extended in
\cite{KumarAssembly:2011} to assembly an infrastructure. A control law
for autonomous landing of an aerobatic airplane on a vertical
surface is proposed in \cite{Stanford_Piercing:2010} and
\cite{Frank.McGrew:07}.  In \cite{IPEK_RAM:2010}, a quadrotor
helicopter is employed to clean a surface while hovering, where an
additional propeller is employed to counteract contact forces while
maintaining the stability of the vehicle.  In
\cite{AUTOMATICA2011_FlyingRobot}, the physical interaction between a
ducted-fan aerial vehicle and the environment is analyzed. The
approach considers to switch the control law in order to take into
account for possible constraints deriving from the presence of
contacts.  Aerial grasping using an autonomous helicopter endowed with
a gripper is considered in \cite{yale} and
\cite{PoundsDollar}. In this case, the analysis focuses on the
stability of the vehicle during the interaction with a compliant
environment. A prototype of miniature aerial manipulator has been
proposed in \cite{Oh2011MMUAV}. In our previous work, we developed a miniature dexterous manipulation system for aerial inspection \cite{keemink2012}, which has been exploited together with a quadrotor UAV for interaction control \cite{matteo} and force regulation \cite{scholten2013}.

In this paper, we present the design, simulation and experimental
validation of a control architecture for a flying hand, which consists of a quadrotor UAV, a robotic
manipulator and a gripper. The overall system can be controlled to have three different operative states: free flight,
dock on the object attached to a vertical surface, fly away with
the grasped object. The stability of the control architecture is achieved by
considering standard passivity-based design techniques, where
the stability of a desired equilibrium point is obtained by shaping
the energy function of the system to have a desired minimum
(\textit{energy-shaping}), and then by dissipating energy to
asymptotically converge to it (\textit{damping-injection}).

The paper is organized as follows. In Section \ref{sec:dynamics}, the
overall flying hand is presented together with its dynamic model. In Section
\ref{sec:control}, we propose the control strategy, which is
validated in both simulations and experimental tests in Section \ref{sec:simulation}. Finally, conclusions are drawn in Section \ref{sec:conclusions}.

\section{SYSTEM DYNAMICS}
\label{sec:dynamics}
In order to design the control law and analyze the stability of the
flying hand while grasping an object, the dynamic model of the system
is required. First, we discuss the structure of the complete
system and, secondly, each part is modelled individually.

\begin{figure}[t]
  \centering
    \includegraphics[width=\columnwidth]{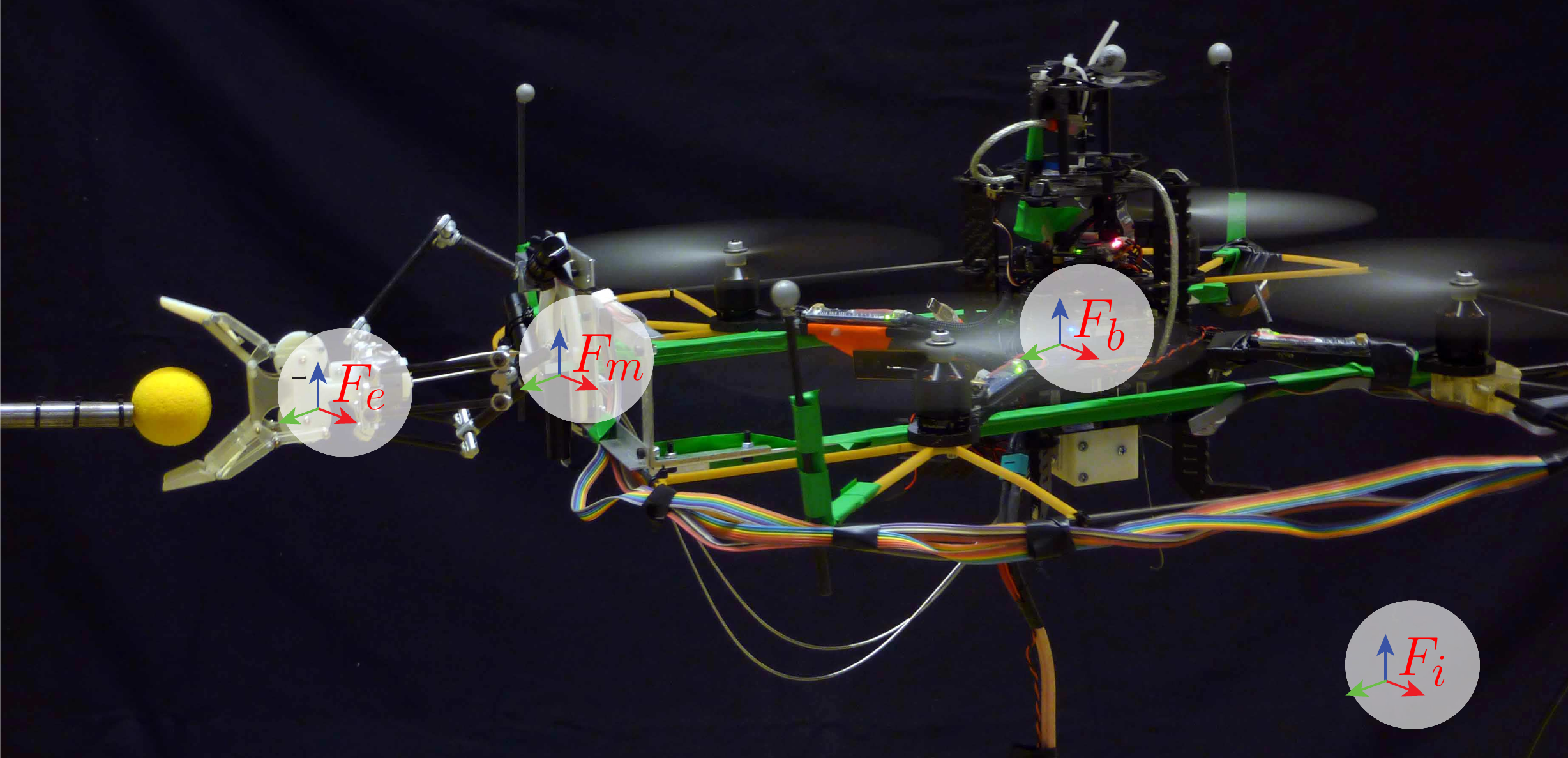}
  \caption{The overall system and its reference frames.}
  \label{fig:notation}
\end{figure}

\subsection{System Overview}
\begin{figure}[t]
  \centering

  \includegraphics[width=\columnwidth]{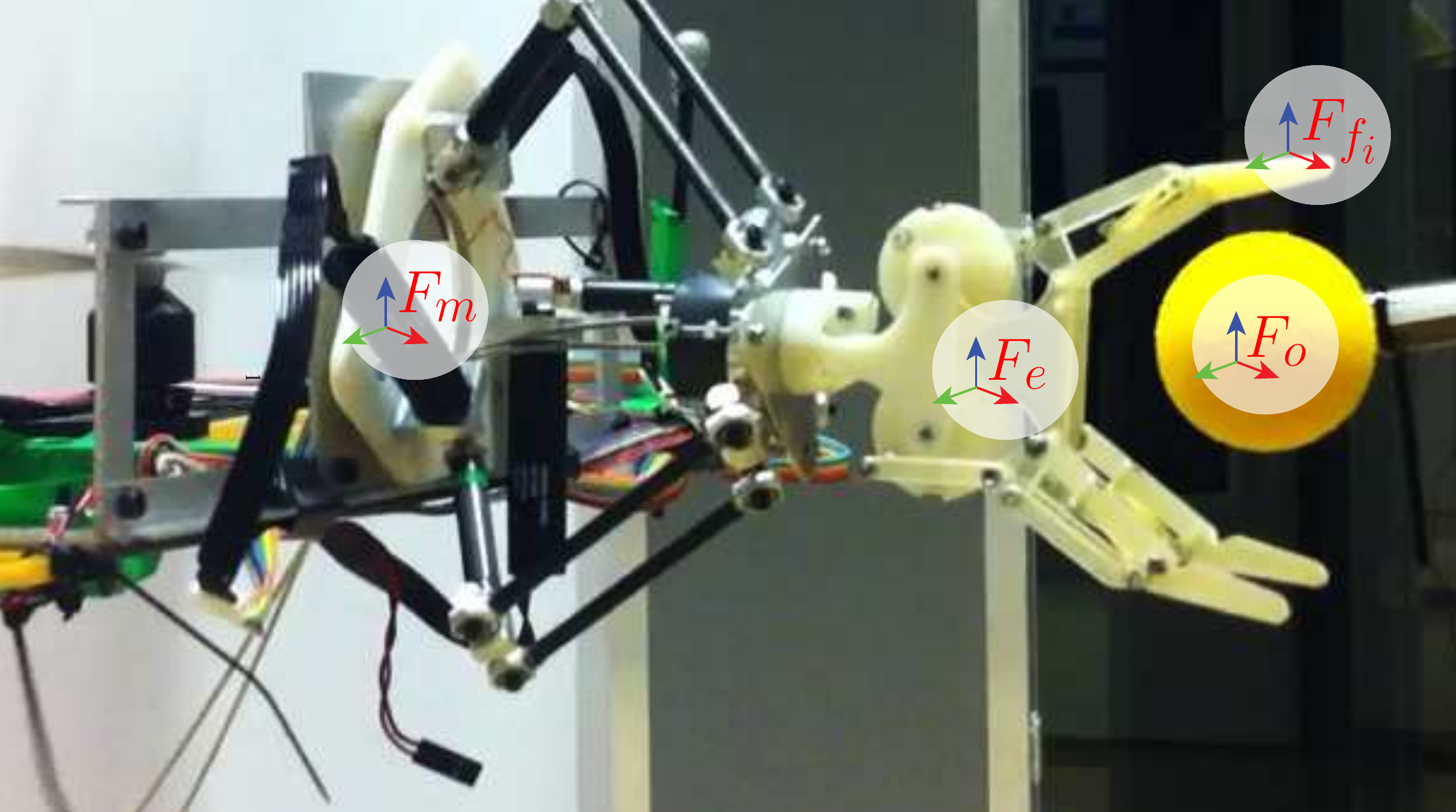}
  \caption{The robotic manipulator, the gripper, the object and
          their reference frames.}
  \label{fig:notation2}
\end{figure}

The complete system is composed of three main parts, as shown in Figure~\ref{fig:notation}. The first is the
underactuated quadrotor UAV. The second part is the robotic manipulator, whose design is based on a delta parallel kinematic structure,
as presented in \cite{keemink2012}. It enables Cartesian
movement of its end-effector in the workspace and it is attached to one side of the quadrotor UAV. 
Finally, the third part is the underactuated gripper, whose design is
inspired by the work presented in \cite{underactuated_hands}. The
mechanical structure of the gripper consists of three fingers, with
two phalanges each, and is actuated by one single motor. The
construction of the gripper is such that form closure is
guaranteed. The robotic manipulator and the gripper are
shown in Figure~\ref{fig:notation2}.

\subsection{Notation}
Before proceeding with the description of the system dynamics, all
used symbols are briefly explained for clarity. With reference to Figures
\ref{fig:notation} and \ref{fig:notation2}, the kinematic
notation is:
\begin{itemize}
\item $F_i$, $F_b$, $F_m$, $F_e$, $F_{f_i}$, and $F_o$ represent the inertial
  frame, the body frame of the UAV fixed at its centre of gravity
  (c.g.), the base frame of the robotic manipulator, the base frame of
  the palm of the gripper (coincident to the base frame of the end
  effector of the manipulator), the frame at the contact points on the
  fingers of the gripper, and the object frame;
\item $p_\alpha^\beta = [x_\alpha^\beta, y_\alpha^\beta, z_\alpha^\beta]^T$ and $R_\alpha^\beta \in
  \mathbb{R}^{3\times3}$ the position and rotation matrix of a generic frame $F_\alpha$
   with respect to a generic frame $F_\beta$.
\end{itemize}
The dynamic notation is:
\begin{itemize}
\item $g$ the gravitational acceleration;
\item $m_\textit{uav}$, $J_\textit{uav}$ are the UAV's mass and
  inertia matrix;
\item $m_\textit{man}$ and $m_\textit{obj}$ are the mass of the robotic manipulator (including the palm on the gripper) and of the object, respectively;
\item $f_{p}^b \in \mathbb{R}$ the total thrust on the vehicle
  generated by its propellers, $f_\textit{man}^m \in \mathbb{R}$ the
  force the robotic manipulator and the UAV exert on each other at the
  robotic manipulator base;
\item $M_\textit{gy} \in \mathbb{R}^3$ the moment vector due to the
  gyration effects of the propellers, $M_p^b = [M_x, M_y, M_z]^T$ the
  control torque of the vehicle, $M_\textit{man}^m \in \mathbb{R}^3$
  the reaction torque the robotic manipulator and the UAV exchange at
  the robotic manipulator base;
\item $f_{I_m}^m, M_{I_m}^m \in \mathbb{R}^3$ the 
  forces and moments due to the absolute motion of the
  robotic manipulator in $F_m$;
\item $f_{I_h}^e, M_{I_h}^e \in \mathbb{R}^3$ the 
  forces and moments due to the absolute motion of the
  robotic manipulator in $F_e$;
\item $f_{I_{\textit{obj}}}^o, M_{I_{\textit{obj}}}^o \in \mathbb{R}^3$ the 
  forces and moments due to the absolute motion of the
  object in $F_o$;
\item $f_h^e, M_h^e \in \mathbb{R}^3$ the force and moment 
  the gripper and the robotic manipulator's end-effector exert on each
  other;
\item $f_\textit{obj}^{o}, M_\textit{obj}^{o} \in \mathbb{R}^3$ the force and the moment 
that make up the total wrench, $w_\textit{obj}$, the object exerts on the gripper's fingers and palm in the object frame, $F_o$.
\end{itemize}

The dynamic model of the complete system is seen as a
cascade of subsystems, interconnected at certain points by means of
localized interaction forces and moments. It is assumed that the
system interacts with the environment by means of the manipulator's end effector
and the phalanges only, i.e. only in $F_e$ and $F_{f_i}$. Furthermore, the
connections between the UAV and the robotic manipulator's base in
$F_m$ as well as between the robotic manipulator's end-effector and
the gripper palm in $F_e$ are assumed to be rigid. The UAV is unconstrained in
$F_b$ and can thus move freely with respect to the
inertial frame, $F_i$.

\subsection{The Quadrotor Dynamics}
The quadrotor is an underactuated system, since it has only four
control inputs $f_i$, i.e. its propellers, and six degrees of freedom
(DoFs). Due to the mechanical design of the quadrotor, a net torque can
be applied in any direction by varying the relative thrust of the
propellers. The mapping between the generated force of each propeller
and the total thrust and torque exerted on the UAV's c.g. is given by
\begin{align*}
\left[
\begin{array}{c}
f_p^b\\
M_x\\
M_y\\
M_z
\end{array}
\right] = \left[
\begin{array}{cccc}
1 & 1 & 1 & 1 \\
0 & -d & 0 & d \\
d & 0 & -d & 0 \\
-c & c & -c & c
\end{array}
\right] \cdot \left[
\begin{array}{c}
f_1 \\
f_2 \\
f_3 \\
f_4
\end{array}
\right]
\end{align*}
where $d$ is the distance between the UAV's c.g. and the centre of the
propeller, and $c$ is the ratio between the propeller reaction torque
and generated thrust. Using this, the full dynamics of the UAV can be
described by
\begin{align}
m_\textit{uav} \cdot \dot{v}^i &= m_\textit{uav} \cdot g \cdot
\hat{z}^i + f_p^b \cdot R_b^i \cdot [0,0,-1]^T\notag + R_b^i \cdot R_m^b \cdot f_\textit{man}^m\notag\\
J_\textit{uav} \cdot \dot{\omega}_b^{b,i} &= -\omega_b^{b,i} \times
J_\textit{uav} \cdot \omega_b^{b,i} + M_\textit{gy} + M_p^b\notag\\
~&\quad+ R_b^i \cdot R_m^b \cdot M_\textit{man}^m + R_b^i (R_m^b \cdot f_\textit{man}^m \times p_m^b)
\label{eq:quad}
\end{align}
where $\omega_b^{b,i}$ and $\dot{\omega}_b^{b,i}$ are the rotational
velocity and acceleration of the UAV's c.g. with respect to $F_i$ expressed
in $F_b$, respectively; $\dot{v}^i$ is the linear acceleration of the
UAV's c.g. expressed in $F_i$.

\subsection{The Robotic Manipulator Dynamics}
The dynamics of the robotic manipulator can be compactly described by
considering the internal and external contributions. The former
includes inertial, Coriolis, centrifugal and gravitational forces/moments as well as the
dynamic influence of the actuators. The latter upholds forces and
moments exchanged at the end effector with the gripper and due to
interaction with the environment. From the equilibrium of forces and
moments, the dynamics of the robotic manipulator in $F_m$ is given
by
\begin{align}
f_\textit{man}^m &= f_{I_m}^m + R_e^m \cdot f_h^e\notag\\
M_\textit{man}^m &= M_{I_m}^m + R_e^m \cdot M_h^e + R_e^m \cdot f_h^e \times p_e^m
\label{eq:delta}
\end{align}

\subsection{The Gripper Dynamics}
Similarly to the robotic
manipulator, the dynamics of the gripper can be compactly described by considering the internal and external contributions. %
Therefore, the dynamics of the
gripper is described by
\begin{align}
f_h^e &= f_{I_h}^e + R_o^e \cdot f_\textit{obj}^o \notag\\
M_h^e &= M_{I_h}^e + R_o^e \cdot M_\textit{obj}^o + R_o^e \cdot f_\textit{obj}^o \times p_o^e 
\label{eq:hand_dyn}
\end{align}

\subsection{The Environment}
The system can be in three different states: free-flight (no
object), dock (on the object) and aerial grasp (with
object). Therefore, both $f_\textit{obj}^o$ and $M_\textit{obj}^o$ in Eq.~\ref{eq:hand_dyn} can have three different interpretations,
depending on which state the system is in. These three states and
their influence on the system are explained hereafter.

\subsubsection{Free-flight state (no object)}
In this state, the UAV is in free flight and no object is grasped. This
results in no external force and moment, i.e.,
\begin{align*}
f_\textit{obj}^{o} &=0 \notag\\
M_\textit{obj}^{o} &=0
\end{align*}

\subsubsection{Dock state (on the object)}
\label{sec:docking}
In this state, the UAV is docked on the object and, more precisely, it is docked by means of the gripper that is grasping the object attached to a vertical surface. The reaction forces and moments are partly due to the contact between the object and the palm of the gripper and partly due to the interaction between the object and the phalanges of the gripper. These forces and moments are described by
\begin{align}
[ f_\textit{obj}^{o}, M_\textit{obj}^{o} ]^T = G \cdot f_c
\label{eq:docked}
\end{align}
where $G$ is the grasp matrix 
and
\begin{align*}
f_c =[ f_c^{f_1}, f_c^{f_2}, f_c^{f_3}, f_c^{f_4}, f_c^{f_5}, f_c^{f_6}, f_c^e ]^T
\end{align*}
is the net vector of the contact forces on the phalanges and the palm. Each element of $f_c$ consists of a normal component, modelled with the Hunt-Crossley model \cite{huntcrossley}, and two tangential friction components.

\subsubsection{Aerial grasp state (with object)}
When the object is detached from the vertical wall, it becomes part of
the complete system. The dynamic contribution of the object is
\begin{align*}
f_\textit{obj}^{o} &=f_{I_{\textit{obj}}}^o \notag\\
M_\textit{obj}^{o} &=M_{I_{\textit{obj}}}^o
\end{align*}

\section{CONTROL}
\label{sec:control}
In this section we present the control architecture, which consists of a cascade of three impedance controllers. Each controller is designed separately for each one of the three subsystems, as described in Section \ref{sec:dynamics}. 


\subsection{The Quadrotor}
Assuming a high attitude control authority like in \cite{matteo}, which
compensates the momenta imposed on the system, the UAV's system dynamics
in Eq.~\ref{eq:quad} can be simplified as
\begin{align}
m_\textit{uav} \cdot \dot{v}^i &= m_\textit{uav} \cdot g \cdot \hat{z}^i + f_p^b \cdot R_b^i \cdot [0,0,-1]^T + R_b^i \cdot R_m^b \cdot f_\textit{man}^m
\label{eq:red_quad}
\end{align}
where $f_p^b \cdot R_b^i \cdot [0,0,-1]^T$ is the only controllable input. 
This is chosen to be
\begin{align}
  f_p^b \cdot R_b^i \cdot [0,0,-1]^T = u_\textit{uav} - m_\textit{uav} \cdot g
  \cdot \hat{z}^i
  \label{eq:fpb}
\end{align}
with $u_\textit{uav} \in \mathbb{R}^3$ as a new
input. Substituting Eq.~\ref{eq:fpb} into Eq.~\ref{eq:red_quad}, it results
\begin{align*}
m_\textit{uav} \cdot \dot{v}^i &= u_\textit{uav} + F_\textit{ext}(t)
\end{align*}
in which $F_\textit{ext}(t) = R_b^i \cdot R_m^b \cdot
f_\textit{man}^m(t)$. In this way, the system resembles a mass driven
by an external force and an input that has still to be designed.

Now let $p_b^{* i}$ be the desired position for the aerial vehicle. By
choosing $u_\textit{uav}$ to be equal to
\begin{align*}
u_\textit{uav} = -K_\textit{uav} (p_b^i-p_b^{* i}) - D_\textit{uav} \dot{p}_b^i - u_{{FF}_{uav}}
\end{align*}
the system is impedance controlled. The gains $K_\textit{uav}$ and $D_\textit{uav}$ can be
chosen according to the desired system bandwidth and relative
damping; $u_{{FF}_{uav}}$ is a feed-forward term that compensates for the gravitational contributions of the robotic manipulator, the gripper and the object, when detached from the wall.

\subsection{The Robotic Manipulator}
In its essence the control of the robotic manipulator is similar to
the control of the UAV. The only difference is the way the inputs map
to the robotic manipulator's end effector position, which is described
by the Jacobian of the delta structure. Note that the high attitude control authority on the UAV compensates for the momenta $M_\textit{man}^m$ that the robotic manipulator and the UAV exchange.

By making $f_{I_m}^m$ explicit in Eq.~\ref{eq:delta}, 
the dynamics of the robotic manipulator can be written as
\begin{align}
  f_\textit{man}^m &= -\eta_\textit{man} + m_\textit{man} \cdot R_e^m \cdot \dot{v}^e + R_e^m
  \cdot f_h^e
\label{eq:delta_mech}
\end{align}
where $\eta_\textit{man}$ contains the gravitational, Coriolis and centrifugal components of the robotic manipulator, $\dot{v}^e$ is the
end effector acceleration and in which, for simplicity, we assumed that the manipulator mass is concentrated on the end-effector. %
%
The controllable input part $f_\textit{man}^m$ is chosen to be
\begin{align}
f_\textit{man}^m = u_\textit{man} - \eta_\textit{man}
\label{eq:delta_control}
\end{align}
with $u_\textit{man} \in \mathbb{R}^3$ as a new input. By
substituting Eq.~\ref{eq:delta_control} in Eq.~\ref{eq:delta_mech}, the following is
obtained
\begin{align}
m_\textit{man} \cdot R_e^m \dot{v}^e &= u_\textit{man} + F_\textit{m,ext}(t)
\end{align}
in which $F_\textit{m,ext}(t) = -R_e^m f_h^e(t)$.

Again the dynamics of this subsystem resembles a mass driven by an
external force. 
Let a desired virtual point be represented by $p_e^{* m}$, if $u_\textit{man}$ is defined as
\begin{align}
u_\textit{man} = -K_\textit{man} (p_e^m-p_e^{* m}) - D_\textit{man} \cdot \dot{p}_e^m - u_{{FF}_{\textit{man}}}
\end{align}
the system is impedance controlled. The gains $K_\textit{man}$ and $D_\textit{man}$ can be
chosen according to the desired system behavior; $u_{{FF}_{\textit{man}}}$ is a feed-forward term that compensates for the gravitational contributions of the gripper and the object, when detached from the wall.

\subsection{The Gripper}
In order to have the overall system consisting of impedance controlled subsystems, we design the control for the gripper in a similar manner as the previous subsystems. Before that, it should be noted that the only interesting state from a control point of view is the dock state, since in both other states the gripper is either idle in its open state or idle in a closed state. Note that the high attitude control authority on the UAV compensates for the momenta $M_\textit{h}^e$ that the gripper and the robotic manipulator exchange.

From Eq.~\ref{eq:hand_dyn} in the dock state, it results in
\begin{align}
f_h^e &= - \eta_\textit{h} + f_\textit{phal}+ R_o^e \cdot f_\textit{obj}^o
\label{eq:fhe}
\end{align}
where $\eta_\textit{h}$ contains the gravitational, Coriolis and centrifugal components of the gripper, $f_\textit{phal}$ are the inertial forces of the phalanges. Note that the mass of the palm is not considered here since it is included in the end-effector. 

The controllable input part $f_h^e$ is chosen to be
\begin{align}
f_h^e &= u_\textit{h} - \eta_\textit{h}
\label{eq:fhe2}
\end{align}
where $u_\textit{h} \in \mathbb{R}$ is a new input. By substituting Eq. \ref{eq:fhe2} in
Eq.~\ref{eq:fhe}, it follows that
\begin{align*}
f_\textit{phal} = u_\textit{h} + F_\textit{h,ext}(t)
\end{align*}
with $F_\textit{h,ext}(t) = - R_o^e \cdot f_\textit{grasp}^o $. The gripper is impedance controlled if the input $u_\textit{h}$ is chosen as
\begin{align*}
u_\textit{h} = -K_\textit{h} (p_f^e-p_f^{* e}) - D_\textit{h} \dot{p}_f^e - u_{FF_{\textit{h}}}
\end{align*}
where $p_f^{* e}$ is the desired position of the gripper's fingers, $K_\textit{h}$ and $D_\textit{h}$ are properly chosen gains, and $u_{FF_{\textit{h}}}$ is a feed-forward term that compensates for the gravitational contributions of the object, when detached from the wall.

Note that as the gripper is an underactuated system, there is only one actuator to actuate three fingers with two phalanges each. This implies that the pseudoinverse of the grasp matrix $G$ should be computed to calculate $p_f^e$ and $\dot{p}_f^e$.


\subsection{Stability Analysis}
\label{sec:stability}
The three subsystems, as described in the previous subsections, have the same generalized closed
loop system, i.e.,
\begin{align}
m \ddot{p}^j + D \dot{p}^j + K \cdot (p^j - p^{*j}) = d
\label{eq:closed_loop}
\end{align}
where $j$ denotes the frame that corresponds to
the subsystem. Therefore a single stability analysis will suffice. In
fact the described system turns out to be \emph{output strictly
  passive} \cite{prentice} by choosing input $d$, output $\dot{p}^j$ and
storage function $V(\dot{p}^j,p^j) = T(\dot{p}^j) + P(p^j)$,
where $T(\dot{p}^j)$ is the kinetic energy and $P(p^j)$ is the potential energy, which has a minimum at the desired position $p^{*j}$. As shown in
\cite{prentice}[Lemma 6.7] the above property of \emph{output strict
  passivity} can be linked to zero-input asymptotic stability via
zero-state observability and can be shown to hold for the generalized
dynamics of the subsystems in Eq.~\ref{eq:closed_loop}. This means that all the subsystems asymptotically reach the desired
set-points, denoted $p^{*j}$ in Eq.~\ref{eq:closed_loop}, provided
their input forces are zero. Due to the cascaded nature of the subsystems and the fact that all of
the subsystems are asymptotically stable, the overall system is also
asymptotically stable \cite{hogan}.

\section{RESULTS}
\label{sec:simulation}
In this section, we provide simulations and experimental tests that validate the proposed control strategy.

\subsection{Simulations}
To simulate the whole system, the model is implemented in the
simulation package 20-sim (Controllab Products B.V., The Netherlands). By simulating the model, it
can be shown that stable flight and stable interaction can be
achieved.

\subsubsection{Quadrotor path tracking}
\begin{figure}[t]
\centering
  \includegraphics[width=\columnwidth]{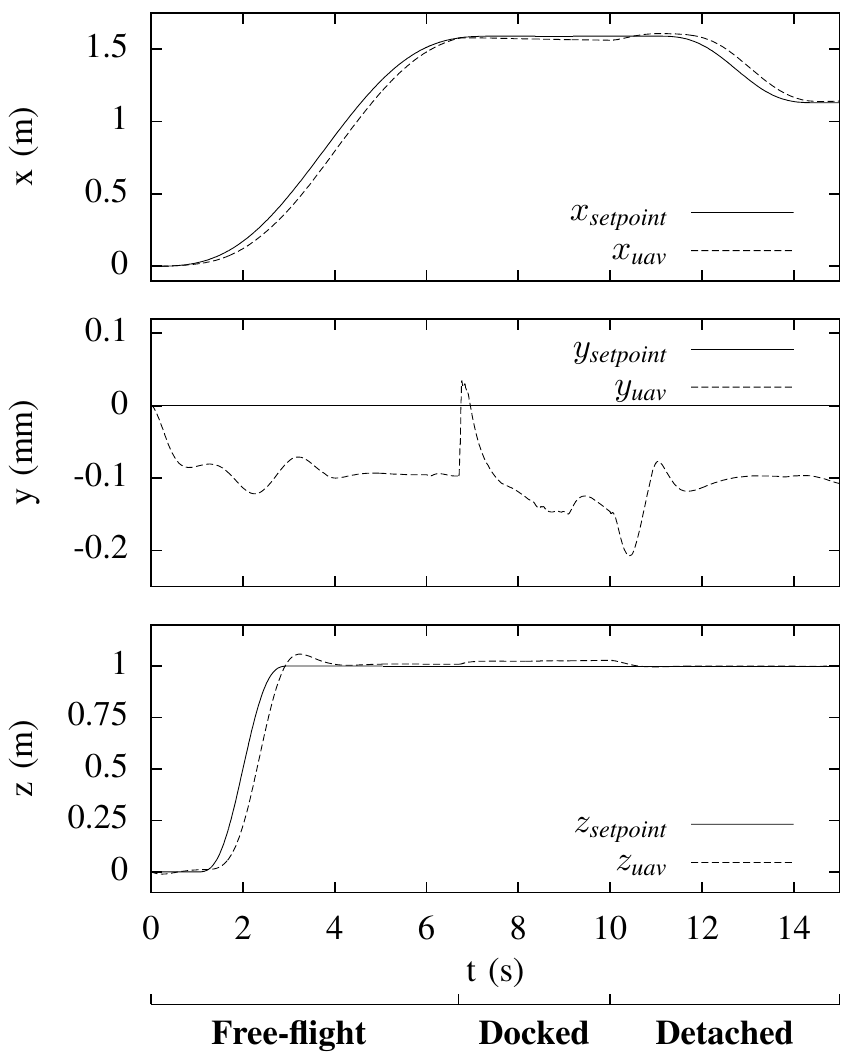}
  \caption{Simulation results - Path tracking of the UAV with respect to the inertial frame $F_i$.}
  \label{fig:track_uav}
\end{figure}

Figure~\ref{fig:track_uav} shows the simulation results for the quadrotor UAV
while tracking a certain path. The three operating states are shown sequentially.

At the beginning, the UAV is in \emph{free-flight state (no object)}. Upon
contact with the object at $t=6.7$ s, the system is in \emph{dock state (on the object)}. When the object is fully grasped and the UAV
flies away from the wall at $t=11$ s, the object is detached from the
wall and becomes a part of the complete system. The system is
finally in its \emph{aerial grasp state}. 

The quadrotor is capable of
tracking the path in all the states. Some tracking errors are present,
due to impedance control, and some larger fluctuations can be seen
when the system is in contact with the environment, even if stability
is guaranteed. The tracking error converges to zero when the
setpoint is constant for a sufficient amount of time. This is in
accordance with the stability analysis in Section~\ref{sec:stability}.

\subsubsection{Robotic manipulator path tracking}
\begin{figure}[t]
  \centering
  \includegraphics[width=\columnwidth]{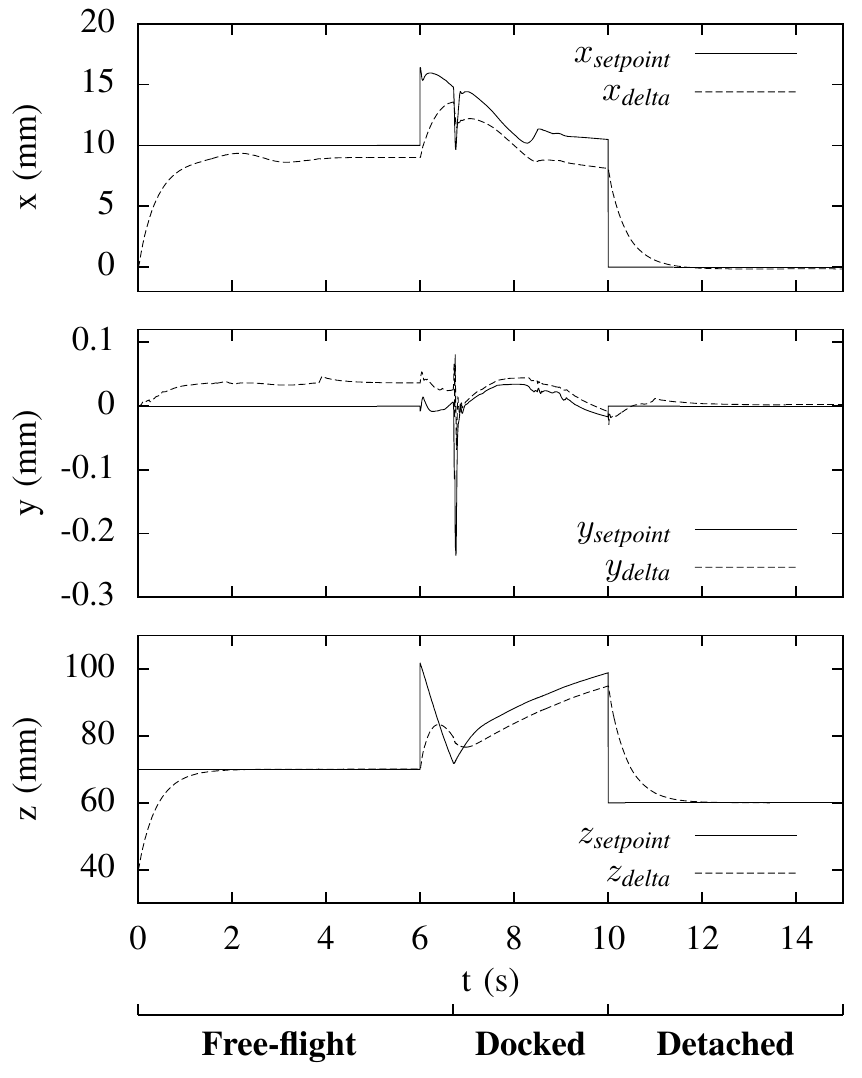}
  \caption{Simulation results - Path tracking of robotic manipulator with respect to manipulator base frame $F_m$.}
  \label{fig:track_delta}
\end{figure}

To asses the correct functioning of the robotic manipulator,
the tracking capability with respect to the manipulator base frame,
$F_m$, is shown in Figure~\ref{fig:track_delta}.

Note that object tracking is disabled for $t < 6$ s as there is no
need to track any object yet, and enabling object tracking would only
result in unnecessary disturbances on the system. At $t = 10$ s, the
object tracking is disabled again as this is no longer useful when the
object is detached from the wall.

The tracking
capabilities of the robotic manipulator prove to be sufficient to retain a
stable system. Note that larger fluctuations occur when the system is
in contact with the environment, but the tracking and the overall
stability are still guaranteed. The spikes at the time of impact are
due to the displacement of the UAV on impact, as shown in the middle
plot in Figure \ref{fig:track_uav}. The sudden change in
$x_\textit{setpoint}$, which is the height with respect to $F_m$, can
be caused by the contact friction and by the fact that the quadrotor
is not yet stabilized.

\begin{figure}[t]
  \centering
  \includegraphics[width=\columnwidth]{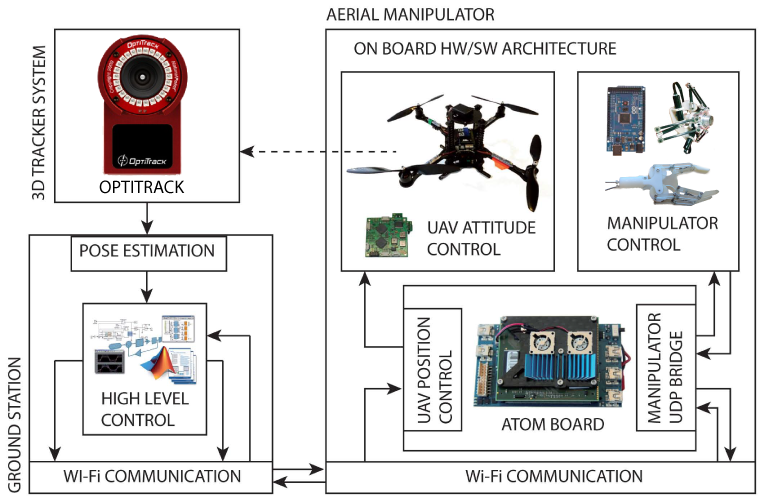}
  \caption{The overall system used for experiments.}
  \label{fig:total_scheme}
\end{figure}

\subsection{Experimental setup}
To verify the proposed control strategy, a test setup has
been build together with a software architecture, as sketched in 
Figure \ref{fig:total_scheme}. The different modules are:
\begin{itemize}
\item Aerial vehicle: a Pelican quadrotor (Ascending Technologies, Germany), with  payload capability of $500$ g.
\item Robotic manipulator: a $3$ DoFs Cartesian delta robotic
  manipulator \cite{keemink2012}. The control of this
  system runs on an Arduino ATMega2560
  and it receives setpoints by the ground station by means of a WiFi
  data link.
\item Gripper: a 3D printed gripper system consisting of three fingers
  with two phalanges each. The system is actuated by one motor
  attached to the aerial vehicle and connected via bowden cables to
  the gripper in order to minimize the UAV's inertia. The motor itself
  is controlled by the microcontroller on the Arduino.
\item External positioning system: the positions of the subsystems in
  the inertial frame, $F_i$, are estimated by an external optical
  tracking system OptiTrack Flex 13 (NaturalPoint, Inc., USA).
\item Ground station: the high-level controller of the system runs on Ubuntu Linux 10.10. It generates trajectories and estimates positions, based on the input of
  the OptiTrack. The communication
  between the different software modules is performed using the Robot
  Operating System ROS. Setpoints generated for
  both the robotic manipulator and the gripper are relayed to the
  integrated microcontroller via WiFi (802.11n
  standard).
\end{itemize}

\subsection{Experiment validation}
In the experiments, the robotic manipulator approaches
the object in the inertial frame. To achieve that, the
desired pose of the robotic manipulator is derived by using the
measured UAV's position and the desired point on the vertical
wall. Once the UAV is in front of the object, the gripper grasps the
object, detaches it from the wall and flies away.

\begin{figure}[t]
  \centering
   \includegraphics[width=0.9\columnwidth]{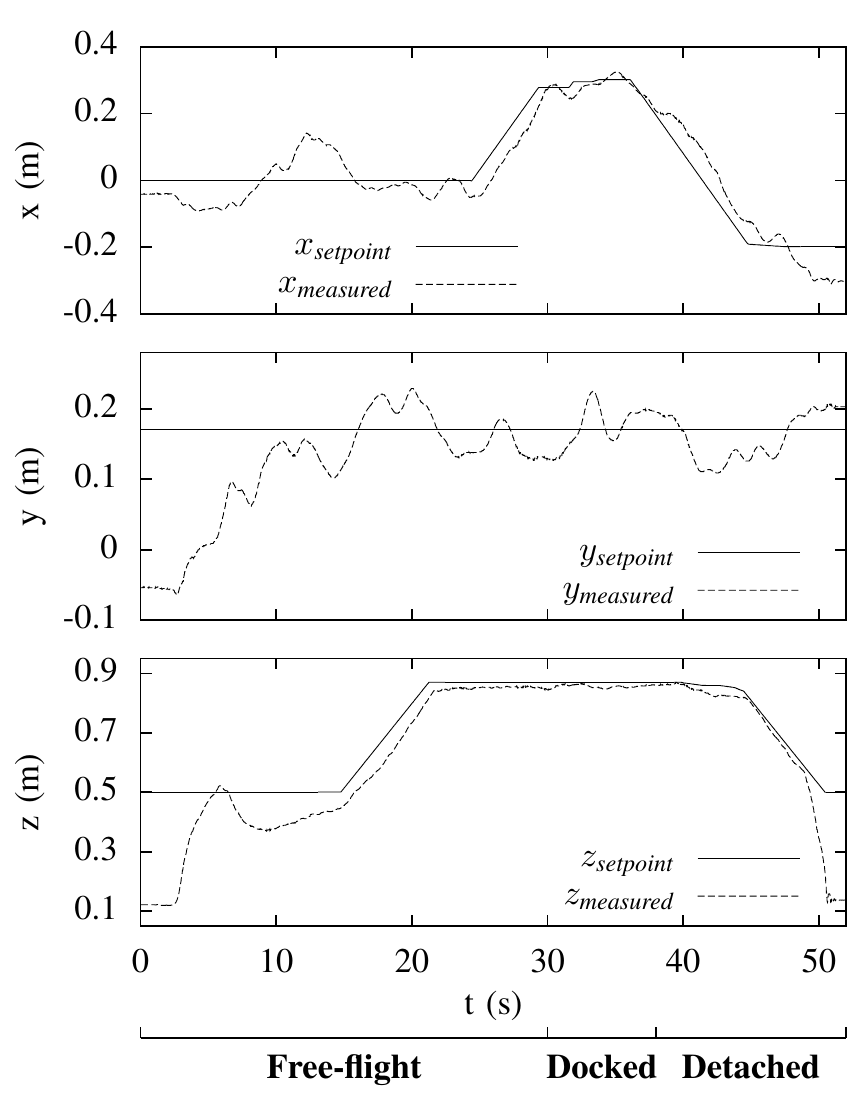}
  \caption{Experimental results - Path tracking of the UAV with respect to the inertial
    frame, $F_i$. $x$ denotes the forward direction, $y$ the sideward
    direction and $z$ the height of the UAV.}
  \label{fig:real_UAV}
\end{figure}

Figure \ref{fig:real_UAV} shows that the UAV is capable of tracking the
given setpoints in all three states. The UAV starts in
free-flight and, at approximately $30$ s, the object is grasped. Few seconds later the object is removed from the vertical wall
leaving the UAV again in free-flight mode. 

\begin{figure}[htp]
  \centering
   \includegraphics[width=0.9\columnwidth]{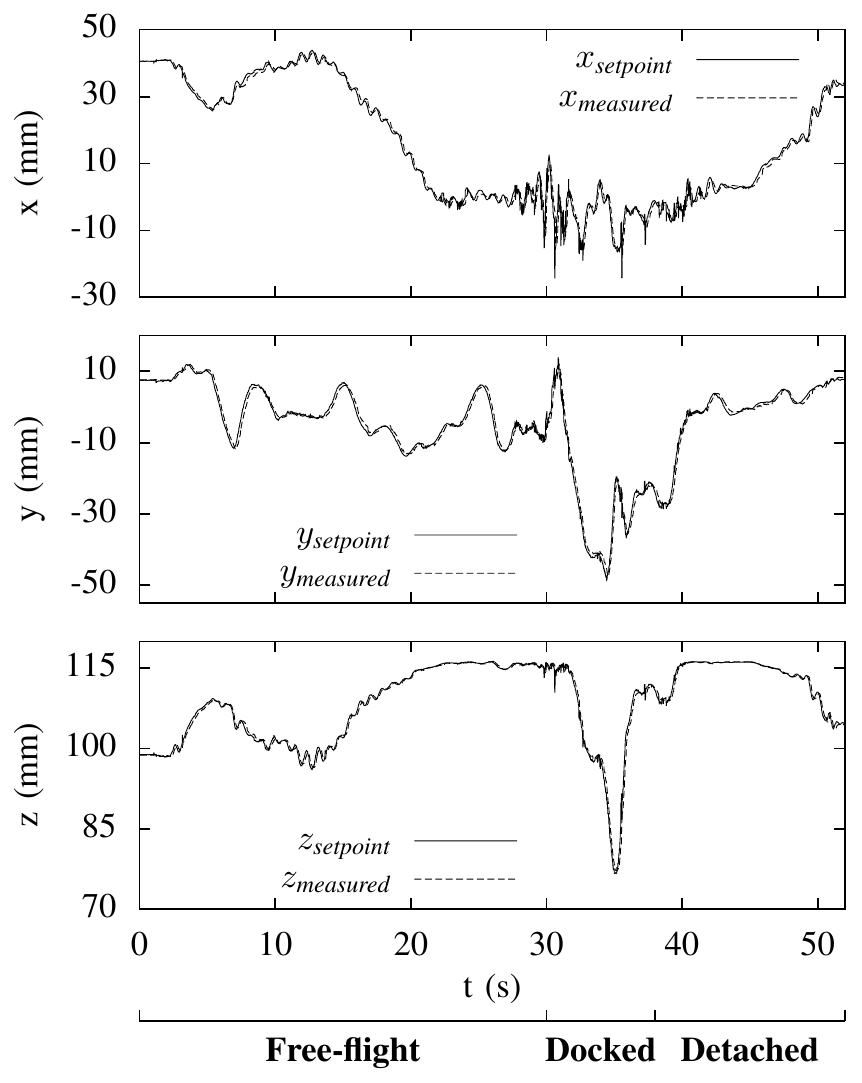}
  \caption{Experimental results - Path tracking of the robotic manipulator's end effector with respect to $F_b$. $x$ denotes the height of the manipulator, $y$ denotes the sideways movement and $z$ denotes the forward motion.}
  \label{fig:real_ee}
\end{figure}

During the experiment the robotic manipulator tracks its setpoint. This can clearly be seen in Figure \ref{fig:real_ee} since the z- and x-position are almost in their outmost position (i.e. $120$ mm for z and $50$ mm for x). At $t=30$ s, the inertial position tracking is visible as the
peaks in both the y- and z-setpoints. After grasping, the robotic
manipulator keeps on tracking the given setpoint which causes it to
move outward again, but with a grasped object. Good tracking
performance and stability are also shown in the accompanying video.

\section{Conclusions}
\label{sec:conclusions}
In this paper, we presented the design of a control architecture for a
flying hand. The overall system consists of a quadrotor UAV, a robotic manipulator and a gripper, which is grasping an
object fixed on a vertical wall. The control strategy is based on
passivity-based techniques and has been shown to guarantee the
asymptotic stability of the system in both simulations and in
experimental tests.

\bibliographystyle{IEEEtran}
\bibliography{references}

\end{document}